\def\year{2019}\relax
\documentclass[letterpaper]{article} 
\usepackage{aaai19}  
\usepackage{times}  
\usepackage{helvet}  
\usepackage{courier}  
\usepackage{url}  
\usepackage{graphicx}  
\usepackage{latexsym}
\usepackage{amsmath}
\usepackage{amsfonts}
\usepackage{todonotes}
\usepackage{textcomp}
\usepackage{tabu}
\usepackage{multirow}
\usepackage{booktabs}
\usepackage{algorithm}
\usepackage[noend]{algpseudocode}
\usepackage[normalem]{ulem}
\usepackage{nicefrac}
\usepackage{csquotes}
\usepackage{changepage}

\DeclareMathOperator*{\argmax}{\arg\!\max}

\def\BState{\State\hskip-\ALG@thistlm}
\newcommand{\seqtoseq}{\textsc{Seq2Seq}}

\begin{document}
\title{Translating Natural Language to SQL using Pointer-Generator Networks and\\How Decoding Order Matters}
\author{
Denis Lukovnikov\textsuperscript{1}, Nilesh Chakraborty\textsuperscript{1}, Jens Lehmann\textsuperscript{1}, and Asja Fischer\textsuperscript{2}\\
\textsuperscript{1}{University of Bonn, Bonn, Germany}\\
\textsuperscript{2}{Ruhr University Bochum, Bochum, Germany}\\
}

\newcommand{\citet}[1]
{\citeauthor{#1}~\shortcite{#1}}
\newcommand{\citep}{\cite}
\newcommand{\citealp}[1]
{\citeauthor{#1}~\citeyear{#1}}

\newcommand{\wikisql}{\textsc{WikiSQL}}

\newcommand{\emb}[1]{\mathbf{#1}}
\newcommand{\enc}[1]{\mathbf{#1}}
\newcommand{\voc}[1]{\mathcal{#1}}
\newcommand{\vectr}[1]{\emb{#1}}		
\newcommand{\outvoc}[1]{\texttt{#1}}

\newcommand{\ourname}{PtrGen-SQL}

\newcommand{\jl}[1]{#1}
\newcommand{\af}[1]{#1}
\newcommand{\dl}[1]{#1}
\newcommand{\nc}[1]{#1}

\nocopyright
\maketitle

\begin{abstract}
Translating natural language to SQL queries for table-based question answering is a challenging problem and has received significant attention from the research community. 
In this work, we extend a pointer-generator and investigate the order-matters problem in semantic parsing for SQL.
Even though our model is a straightforward extension of a general-purpose pointer-generator, it outperforms early works for WikiSQL and remains competitive to concurrently introduced, more complex models.
Moreover, we provide a deeper investigation of the potential order-matters problem that could arise due to having multiple correct decoding paths, and investigate the use of REINFORCE as well as a dynamic oracle in this context.
\footnote{This is an updated version of our previous anonymous version (\url{https://openreview.net/forum?id=HJMoLws2z}) from May 2018.}
\end{abstract}

\section{Introduction}
Semantic parsing, the task of converting Natural Language (NL) utterances to their representation in a formal language,
is a fundamental problem in Natural Language Processing (NLP) and has important applications in Question Answering (QA) over structured data and robot instruction.

In this work, we focus on QA over tabular data, which attracted significant research efforts~\cite{seq2sql,sqlnet,yu2018,huang2018,haug2018,wang2018a,wang2018b,incsql,krishnamurthy2017,iyyer2017,wikitableqa}.
In this task, given a NL  question and a table, the system must generate a query that will retrieve the correct answers for the question from the given table.

The model we use in this paper is a straightforward extension of pointer-generators, and yet outperforms early works and compares well against concurrently developed models. Concretely, we add simple LSTM-based column encoders, skip connections and constrained decoding, as elaborated later in the paper.

In translating NL questions to SQL queries, as in many semantic parsing tasks, target queries can contain unordered elements, resulting in multiple valid decoding paths. It has been shown by~\citet{ordermatters} that in the case of multiple valid decoding paths, the order of sequences provided for supervision can affect the accuracy of \seqtoseq{} models.
We provide a deeper investigation of the potential order-matters problem in translating NL to SQL that has been raised by previous work. In this context, we also investigate training with a dynamic oracle~\citep{goldbergnivre} as well as training with REINFORCE, both of which explore different possible linearizations of the target queries, and show that the use of a dynamic oracle can be beneficial when the original supervision sequences are ordered inconsistently.

In the following, we first introduce the problem, then describe our model and the training procedure, present \af{an experimental analysis}, and conclude with a comparison to related work.

\section{Queries, Trees and Linearizations}
\label{sec:example}

As an illustration \af{of table-based QA}, 
consider the \af{natural language} question 
\begin{displayquote}
``\textit{How much L1 Cache can we get with an FSB speed of 800MHz and a clock speed of 1.2GHz?}''
\end{displayquote}	
\dl{This question} should be mapped to the following \af{SQL} query\\
\begin{tabular}{r l}
\outvoc{SELECT} & \outvoc{L1\_Cache}\\
\outvoc{WHERE } & \outvoc{FSB\_Speed = 800} \textit{(Mhz)} \\
\outvoc{AND}    & \outvoc{Clock\_Speed = 1.0} \textit{(GHz)}\\
\end{tabular}\\
which will be executed over a table listing processors, the sizes of their caches, their clocks speeds etc.
\dl{In the query representation format we use, the example SQL query will be represented as the following sequence of output tokens: }
\dl{\begin{quote}\outvoc{SELECT L1\_Cache AGG0 WHERE COND FSB\_Speed OP0 VAL \textit{800} ENDVAL COND Clock\_Speed OP0 VAL \textit{1.0} ENDVAL} \enspace , \end{quote}}
\noindent \dl{where \outvoc{AGG0} is a dummy ``no aggregator'' token that is used to indicate that no real aggregator should be applied and \outvoc{OP0} is the \outvoc{=} (equality) operator.  Other aggregators, like \outvoc{SUM} and \outvoc{COUNT}, and other operators, like \outvoc{\textless} (less than) are also available.}

\begin{figure}[h]
\includegraphics[width=0.9\linewidth]{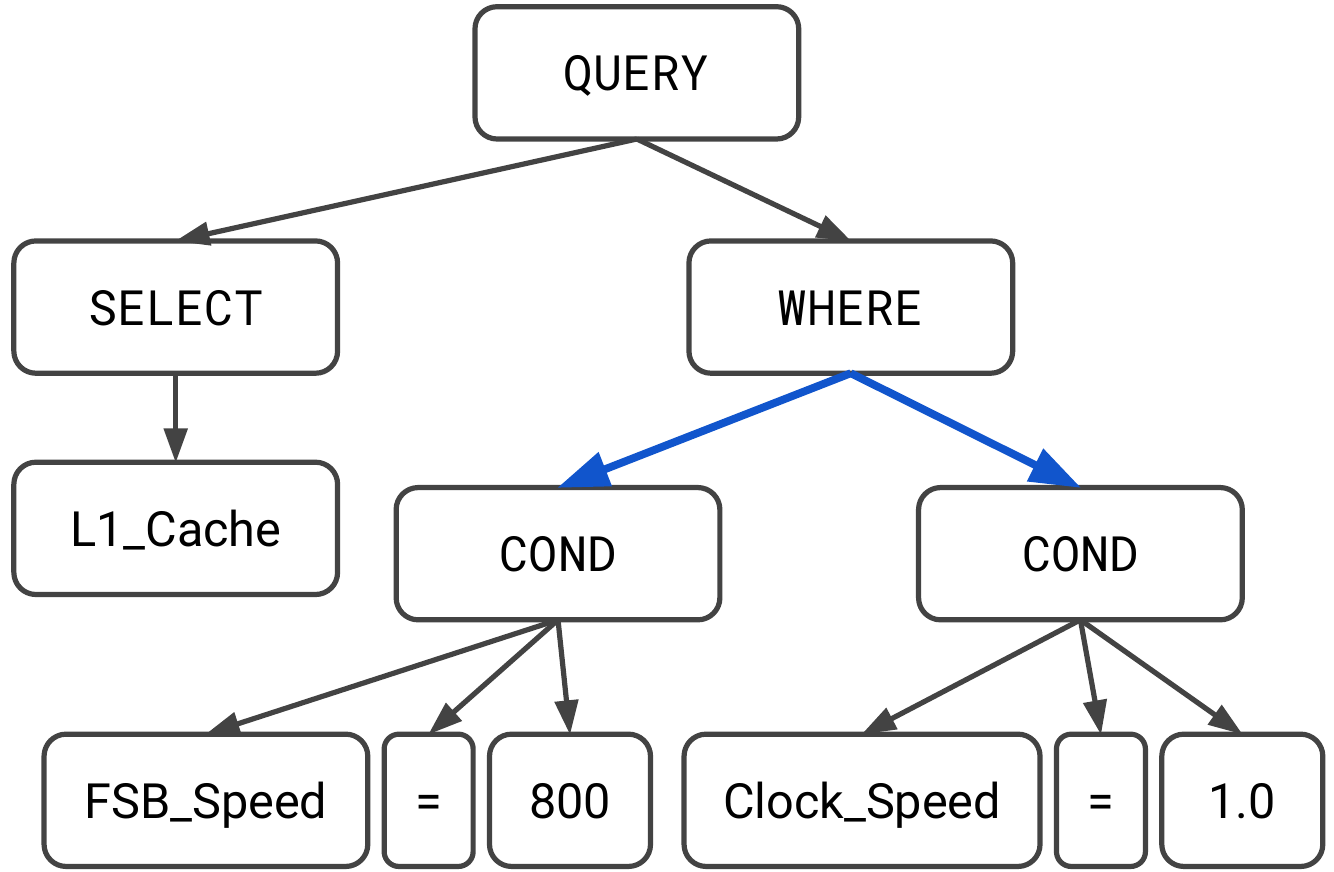}
\centering
\caption{Example of a query tree. The blue arrows indicate unordered children.}
\label{fig:tree}
\end{figure}

\begin{figure}[h]
\includegraphics[width=\linewidth]{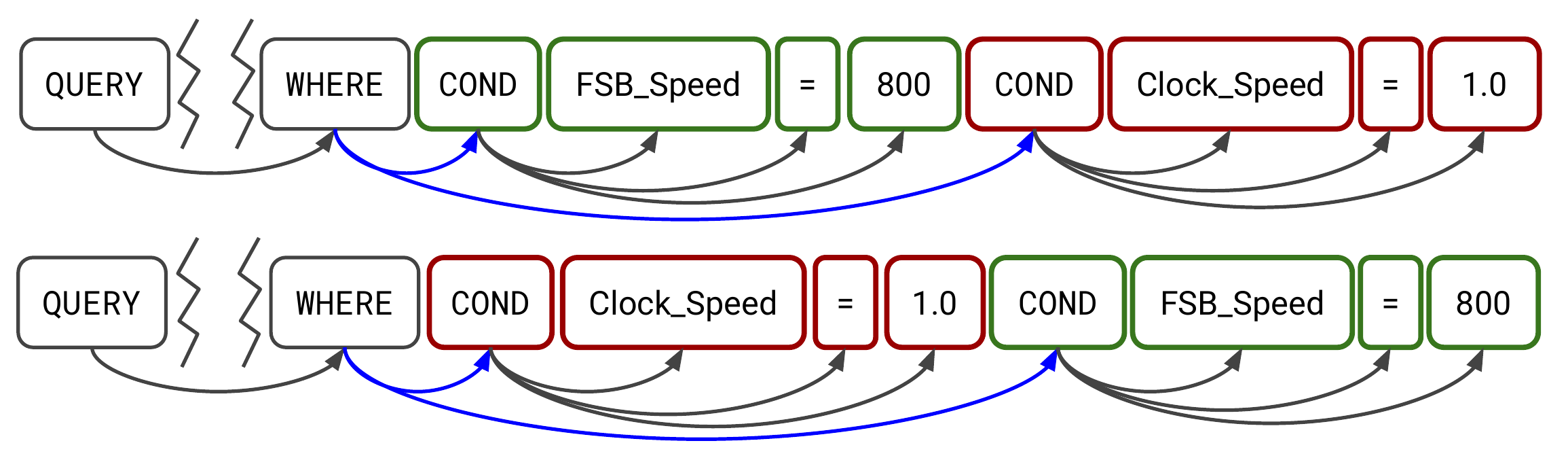}
\centering
\caption{Two valid linearizations of the example query tree in Figure~\ref{fig:tree}.}
\label{fig:linearizations}
\end{figure}

\af{As illustrated in Figure~\ref{fig:tree}, the SQL}   
query can \af{also} be represented as a tree where the root node has two children: \outvoc{SELECT} and \outvoc{WHERE}.
Note that the order of the two conditions appearing in the \outvoc{WHERE} clause is arbitrary and does not have any impact on the meaning of the query or the execution results.
Trees containing such unordered nodes can be linearized into a sequence in different, equally valid, ways ("FSB Speed" first or "Clock Speed" first in the example, \af{as illsustrated in Figure~\ref{fig:linearizations}.}). We refer to the linearization where the original order as given in the data set is preserved as the \textit{original linearization}.

\section{Model}
\label{sec:model}
We start from a sequence-to-sequence model with attention and extend the embedding and output layers to better suit the task of QA over tabular data. 
In particular, we use on-the-fly~\cite{ontheflyemb} embeddings and output vectors for column tokens and implement a pointer-based~\citep{gu2016,see2017} mechanism 
for copying tokens from the question. 
The resulting model is a Pointer-Generator~\citep{gu2016,see2017} with on-the-fly representations for a subset of its vocabulary.

\subsection{The Seq2Seq Model}
\label{sec:seq2seq}
The general architecture of our model follows the attention-based sequence-to-sequence (\seqtoseq{}) architecture.
The following formally introduces the major parts of our \seqtoseq{} model.
Details about the embedding and the output layers
are further elaborated in later sections.

The \seqtoseq{} model consists of an encoder,
a decoder,
and an attention mechanism.

\subsubsection{\af{Encoder}}
\label{sec:encoding}
We are given a question $Q=[q_0, q_1, \dots , q_N]$
\af{consisting} of \af{NL} tokens $q_i$ from the set $\voc{V}^E$ (i.e., the encoder vocabulary).
The tokens are first passed through an embedding layer that maps every token $q_i$ to its vector representation $\emb{q}_i = W^E \cdot \mathsf{one\_hot}(q_i)$
where $W^E \in \mathbb{R}^{|\voc{V}^E| \times d^{emb}}$ is a learnable weight matrix and $\mathsf{one\_hot(\cdot)}$ maps a token to its one-hot vector. 

Given the token embeddings, a bidirectional multi-layered Long Short-Term Memory (LSTM)~\cite{lstm} encoder produces the hidden state vectors
$[\enc{h^*}_0, \enc{h^*}_1, \dots, \enc{h^*}_N]$ $= \mathsf{BiLSTM}([\emb{q}_0, \emb{q}_1, \dots, \emb{q}_N])$.

The encoder also contains
skip connections
that add 
word embeddings $\emb{q}_i$ to the 
hidden states
$\enc{h^*}_i$.

\subsubsection{\af{Decoder}}
\label{sec:decoder}
The decoder produces a sequence of output tokens $s_t$ from an output vocabulary $\voc{V}^D$ conditioned on the input sequence $Q$. 
It is realized by a uni-directional multi-layered LSTM.
First, the previous output token $s_{t-1}$ is  mapped to its vector representation using the embedding function $\text{EMB}(\cdot)$.
The embeddings are fed to a BiLSTM-based decoder and its output states are used to compute the output probabilities over $\voc{V}^D$ using the output function $\text{OUT}(\cdot)$. $\text{EMB}(\cdot)$ and $\text{OUT}(\cdot)$ are described in the following sections.

\subsubsection{Attention}
\label{sec:attention}
We use attention~\citep{bahdanau} to compute the context vector $\enc{\hat{h}}_t$, that is
\begin{align}
\label{eq:attention:scores} a^{(t)}_i &= \enc{h}_i \cdot \enc{y}_t \enspace , \\
\label{eq:attention:probs} \alpha^{(t)}_i &= \mathsf{softmax}(a^{(t)}_0, \dots, a^{(t)}_i, \dots, a^{(t)}_N)_i \enspace , \\  
\enc{\hat{h}}_t &= \sum^N_{i=0} \alpha^{(t)}_i \enc{h}_i \enspace ,
\end{align}
where $\mathsf{softmax(\cdot)_i}$ denotes the $i$-ith element of the output of the softmax function and $\enc{h}_1, \dots, \enc{h}_N$ are the 
\af{embedding vectors returned by the}
encoder.

\subsection{Embedding Function of the Decoder}
\label{sec:repr:emb}

The whole output vocabulary $\voc{V}^D$ can be grouped in three parts: (1) SQL tokens from $\voc{V}^{\text{SQL}}$, (2) column ids from $\voc{V}^{\text{COL}}$, and (3) input words from  the encoder vocabulary $\voc{V}^E$, \af{that is}, $\voc{V}^D = \voc{V}^{\text{SQL}} \cup \voc{V}^{\text{COL}} \cup \voc{V}^E$. 
In the following paragraphs, we describe how each of the three types of tokens is embedded in the decoder.

\subsubsection{SQL tokens:} These are tokens 
\af{which are} used to represent the structure of the query,
inherent to the formal target language of choice,
such as SQL-specific tokens like \outvoc{SELECT} and \outvoc{WHERE}. 
\af{Since these tokens have a fixed, example-independent meaning,}
\af{they can be represented by their}
\nc{respective} embedding vectors \nc{shared} across all examples.
Thus, the tokens from $\voc{V}^{\text{SQL}}$ are embedded \af{based on a learnable,} randomly initialized
embedding matrix $W^{\text{SQL}}$ 
\af{which is reused for all examples.}

\subsubsection{Column id tokens:} 
\label{sec:repr:emb:cols}
These tokens are used to refer to \nc{specific} columns in the table \nc{that} the question is being asked against. 

Column names may consist of several words, which are first embedded and then fed into a single-layer LSTM. 
The final hidden state of the LSTM is taken as the embedding vector representing the column.
This approach for computing column representations is similar to other works that encode external information to get better representations for rare words~\cite{ontheflyemb,char2word,embdic}.

\subsubsection{Input words:}
To represent input words in the decoder we reuse the vectors from the embedding matrix $W^{E}$, which is also used for encoding the question.

\subsection{Output Layer of the Decoder}
\label{sec:repr:out}
The output layer of the decoder takes the current context 
$\enc{\hat{h}}_t$ and the hidden state $\enc{y}_t$ of the decoder's LSTM and produces probabilities over the output vocabulary $\voc{V}^D$. 
Probabilities over SQL tokens and column id tokens are calculated based on a dedicated linear transformation, as opposed to \af{the probabilities over} input words \af{which rely on a pointer mechanism that enables copying from the input question.}

\subsubsection{Generating scores for SQL tokens and column id tokens}
For the SQL tokens ($\voc{V}^{\text{SQL}}$), the output scores are computed by the linear transformation: $
\vectr{o}^{\text{ SQL}} = U^{\text{SQL}} \cdot [\enc{y}_t, \enc{\hat{h}}_t]$,
where $U^{\text{SQL}} \in \mathbb{R}^{|\voc{V}^{\text{SQL}}| \times d^{out}}$ is a trainable matrix.
For the \textit{column id tokens} ($\voc{V}^{\text{COL}}$), we compute the output scores based on a transformation matrix $U^{\text{COL}}$, \af{holding  dynamically computed encodings of all column ids present in the table of the current example}. For every column id token, we encode the corresponding column name using an LSTM, taking its final state as \af{a (preliminary)} column name encoding $\enc{u}^*$,
similarly to 
\af{what is done in}
the embedding function.
\af{By using}
skip connections 
\nc{we compute}
the average of the word embeddings of the tokens in the column name, \af{$\emb{c}_i$} for $i=1,\dots, K$,
\nc{and add them} to the  \af{preliminary column name encoding $\enc{u}^*$ to obtain the final encoding for the column id}:
\af{\begin{align}
\enc{u} &= \enc{u}^*+  \begin{bmatrix}
					\emb{0} \\
					\frac{1}{K} \sum_i^{K} \emb{c}_i
				\end{bmatrix}  \enspace ,
\end{align}}
where we pad the word embeddings with zeros to match the dimensions of the encoding vector before adding.

The output scores for all column id tokens are then computed by the linear transformation\footnote{Note that the skip connections in both the question encoder and the column name encoder use padding such that the word embeddings are added to the same regions of $[\enc{y}_t, \enc{\hat{h}}_t]$ and $\enc{u}$, respectively, and thus are directly matched.}: $\label{eq:out:cols}\vectr{o}^{\text{ COL}} = U^{\text{COL}} \cdot [\enc{y}_t, \enc{\hat{h}}_t]$.

\subsubsection{\af{Pointer-based copying from the input}}
\label{sec:repr:out:inps}
To enable our model to copy \af{tokens} from the input question, we
follow a pointer-based~\cite{gu2016,see2017} approach to compute \af{output} scores over the words from the question. 
We explore two different copying mechanism, a \textit{shared softmax} approach inspired ~\citet{gu2016}
 \af{and a \textit{point-or-generate} method 
similar to ~\citet{see2017}. }
 The two  copying mechanisms are described in the following.

\paragraph{Point-or-generate:}
First, the concatenated output scores for SQL and column id tokens are turned into probabilities using a softmax
\af{
\begin{align}
p^{\text{GEN}}(S_t|s_{t-1}, \dots, s_0, Q) &= \mathsf{softmax} ([\vectr{o}^{\text{SQL}} ; \vectr{o}^{\text{ COL}} ])\enspace.
\end{align}}

Then we obtain the probabilities over the input vocabulary $\voc{V}^E$ 
\af{based on}
the attention probabilities $\alpha_i^{(t)}$ (Eq.~\ref{eq:attention:probs}) over the \af{question} sequence \af{$Q = [q_0, \dots, q_i, \dots, q_N]$} .
To obtain the pointer probability for a token $q$  
in the question sequence 
we sum over \af{the attention probabilities corresponding to all the} positions \af{of $Q$}
where $q$ occurs, \af{that is}
\begin{equation}
p^{\text{PTR}}(q|s_{t-1}, \dots, s_0, Q) = \sum_{\af{i:}q_i = q} \alpha_i^{(t)}\ \enspace .
\end{equation}
The pointer probabilities for all input tokens $q$ 
$\in \voc{V}^E$ that do not occur in the question $Q$ are set to $0$.

Finally, the two distributions $p^{\text{GEN}}$ and $p^{\text{PTR}}$ are combined \af{into a mixture distribution:} 
\af{
\begin{align}
p(S_t|s_{t-1}, \dots, s_0, Q) &= \gamma p^{\text{PTR}}(S_t|s_{t-1}, \dots, s_0, Q) \\ &+ (1 - \gamma) p^{\text{GEN}}(S_t|s_{t-1}, \dots, s_0, Q) \enspace ,
\end{align}}
where the scalar mixture weight $\gamma \in [0, 1] $ \af{is} 
\af{given by the output of} a two-layer feed-forward 
\af{neural} network, \af{that gets} \dl{$[\enc{y}_t, \enc{\hat{h}}_t]$} as input.

\paragraph{Shared softmax:}

In this approach, 
we re-use the attention scores $a_i^{(t)}$ (Eq.~\ref{eq:attention:scores}) and obtain the \af{output} scores $\vectr{o}^{\text{ E}}$ \af{over the tokens $q \in \voc{V}^E$ from the question as follows: for} every token $q$ that occurs in the question sequence $Q$  \af{the output score is given by} the maximum \af{attention score} over all positions in 
$Q = [q_0, \dots, q_i, \dots, q_N]$ where 
$q$ occurs, i.e.~\af{it is given} by:
\af{
\begin{equation}
\max_{i:q_i = q} a_i \enspace ,
\end{equation}
}
\af{while} the  scores for all input tokens $q \in \voc{V}^E$ that do not occur in the question $Q$ are set to $-\infty$.
The final output probabilities are then computed 
\af{based on a single softmax function that takes the output scores of the whole output vocabulary as input:}
\begin{align}
p(S_t|s_{t-1}, \dots, s_0, Q) &= \mathsf{softmax} ([\vectr{o}^{\text{SQL}} ; \vectr{o}^{\text{ COL}} ; \vectr{o}^{\text{ E}}])	\enspace .
\end{align}

\subsection{Pretrained Embeddings and Rare Words}
\label{sec:aboutwords}
We initialize all NL embedding matrices\footnote{\af{$W^E$ simultaneously used for question word embedding in the encoder and input word embedding in the embedding function of the decoder,
the embedding matrix $W^{CT}$ for words occurring in column names used in the embedding function of the decoder, and its analogue in the output function.}}
using Glove embeddings for words covered by Glove~\cite{glove} 
and use randomly initialized vectors 
for the remaining words.
Whereas randomly initialized word embeddings are trained together with the remaining model parameters, we keep Glove embeddings fixed, since finetuning them led to worse results in our experiments.

We also replace rare words that do not occur in Glove with a rare word \af{representation}
in all embedding matrices.

\subsection{Coherence of decoded logical forms}
\label{sec:model:rules}
The output sequences produced by a unconstrained decoder can be syntactically incorrect and result in execution errors or they can make mistakes against table semantics.
We avoid such mistakes by implementing a constrained decoder that exploits task-specific syntactic and semantic rules\footnote{We use these constraints during prediction only.}.

The grammar behind the produced sequences is simple and the constraints can be implemented easily by keeping track of the previous token and whether we are in the \outvoc{SELECT} or \outvoc{WHERE} clause. For example, after a \outvoc{COND} token (see also Figure~\ref{fig:tree} and Section~\ref{sec:example}), only column id tokens (\outvoc{L1\_Cache}, \outvoc{FSB\_Speed}, \dots) can follow, and after a column id token, only an operator token (\outvoc{OP1}, \outvoc{OP2}, \dots) is allowed if we are currently decoding the \outvoc{WHERE} clause.

\dl{In addition to such syntactic rules}, we take into account the types of columns to restrict the set of aggregators and operators that can follow a column id. In the case of \wikisql{}, there are two column types: \texttt{text} and \texttt{float}. Aggregators like \textsf{average} and operators like \textsf{greater\_than} only apply on \texttt{float}-typed columns and thus are not allowed after \texttt{text} columns.
We also enforce span consistency when copying tokens, leaving only the choice of copying the next token from the input or terminating copying, if the previous action was a copy action.

\section{Training}
\label{sec:training}
We train our models by maximizing the likelihood of correct logical forms given the natural language question.
We experiment with teacher forcing (TF) and a dynamic oracle~\cite{goldbergnivre}. 

Teacher forcing takes the original linearizations of the query trees (as provided in the dataset) and uses it both for supervision and as input to the decoder.
However, in the presence of different correct sequences (as resulting from different correct linearizations of a query tree), teacher forcing can suffer from suboptimal supervision order~\cite{ordermatters}. This might also be the case for semantic parsing, as pointed out by previous works on \wikisql{} ~\cite{seq2sql,sqlnet} and concurrently explored by~\cite{incsql}.

\subsection{Dynamic Oracle}
\label{sec:training:oracle}
Instead of forcing the \af{model to follow the} original decoding sequence, the dynamic oracle enables the exploration of alternative  linearizations of the query tree and is an adaptation of Goldberg and Nivre's~\shortcite{goldbergnivre} \textit{dynamic oracle with spurious ambiguity}.
It is  formally described  in Algorithm~\ref{alg:oracle}, which is invoked at every decoding step $t$ to get a
token $g_t$ \af{(used for supervision)} and a token \af{$x_{t+1}$} (used as input to the decoder in the next time step).
Essentially, the algorithm always picks the best-scored \textit{correct} token \af{for} supervision and uniformly samples one of the correct tokens 
\af{to be used as decoder input in the next time step}, if the \textit{overall} best-scored token (over the whole output vocabulary) \af{does not belong to the correct ones.} 
Thus, the oracle explores alternative paths if the decoder would make a mistake in free-running mode.
\footnote{Teacher forcing can be seen as a static oracle.}

\begin{algorithm}[h]
\caption{Dynamic oracle}
\label{alg:oracle}
\begin{algorithmic}[1]
\Function{GetNextAndGold}{$p_t$, $t$, $x_{\leq t}$}
\State $\mathsf{VNT}_t \gets \mathtt{get\_valid\_next}(t, x_{\leq t})$
\State $x_{t+1} \gets \argmax_{\voc{V}^D} p_t$
\State $g_t \gets \argmax_{\mathsf{VNT}_t} p_t$
\If {$x_{t+1} \notin \mathsf{VNT}_t$}
    \State $x_{t+1} \gets \mathtt{random}(\mathsf{VNT}_t)$
\EndIf
\State \Return $g_t, x_{t+1}$
\EndFunction
\end{algorithmic}
\end{algorithm}

In the algorithm, $p_t$ is the decoder's output distribution over $\voc{V}^D$ at time step $t$. 
The set of valid next tokens $\mathsf{VNT}_t \subset \voc{V}^D$, from which the correct tree can be reached, is returned by the function $\mathtt{get\_valid\_next}(\cdot)$.
The query tree can have nodes with either ordered or unordered children (for example, children of the \texttt{WHERE} clause are unordered).
If we are currently decoding the children of a node with unordered children, all the children that have not been decoded yet are returned as $\mathsf{VNT}_t$. In other cases, $\mathsf{VNT}_t$ contains the next token according to the original sequence order.

\subsection{REINFORCE}
\label{sec:training:reinforce}
The presented oracle is similar to REINFORCE in that it explores alternative paths to generate the same query. 
In contrast to the oracle, REINFORCE samples the next token ($x_{t+1}$) according to the predictive distribution $p_t$ and then \af{uses} the sampled sequence to compute gradients for policy parameters:
\begin{equation}
\label{eq:pg}
\nabla J = \mathbb{E}[\nabla \log(p_t(x_{t+1})) A_t] \enspace 
\end{equation}
In Alg.~\ref{alg:reinforce}, we adapt the oracle into a REINFORCE algorithm with episode reward $A_t$ set to $+1$ if the sampled sequence produces a correct query and $0$ otherwise.
\begin{algorithm}[h]
\caption{REINFORCE-like oracle}
\label{alg:reinforce}
\begin{algorithmic}[1]
\Function{GetNextAndGold}{$p_t$, $t$, $x_{\leq t}$}
\State $\mathsf{VNT}_t \gets \mathtt{get\_valid\_next}(t, x_{\leq t})$
\State $x_{t+1} \sim p_t ; \enspace x_{t+1} \in \mathsf{VNT}_t$ 
\State $g_t \gets x_{t+1}$
\State \Return $g_t, x_{t+1}$
\EndFunction
\end{algorithmic}
\end{algorithm}

\section{Evaluation}
\label{sec:experiments}
To evaluate our approach, we obtain the \wikisql{} dataset by following the instructions on the \wikisql{} website\footnote{\url{http://github.com/salesforce/WikiSQL}}.

\subsection{Dataset}
Each example in the dataset contains a \af{NL} question, its SQL equivalent and the table against which the SQL query should be executed. 
The original training/dev/test splits of \wikisql{} use disjoint sets of tables with different schemas.

For details on the construction of the dataset and how it compares to existing datasets, we refer the reader to the 
\wikisql{} 
paper \cite{seq2sql}.

\subsection{Experimental Setup}
\paragraph{Evaluation:} Similarly to previous works, we report (1) sequence match accuracy ($\text{Acc}_{\text{LF}}$), (2) query match accuracy ($\text{Acc}_{\text{QM}}$) and (3) query execution accuracy ($\text{Acc}_{\text{EX}}$).
Note that while $\text{Acc}_{\text{LF}}$ accepts only the original linearizations of the trees, $\text{Acc}_{\text{QM}}$ and $\text{Acc}_{\text{EX}}$ accept all orderings leading to the same query.

\paragraph{Training details:} 
After a hyperparameter search, we obtained the best results by using two layers both in the encoder and decoder LSTMs, \af{with} $d^{dec} = 600$ in every layer, and $d^{emb} = 300$ \af{hidden neurons, respectively,}
 and applying time-shared dropouts on the inputs of the recurrent layers (dropout rate 0.2) and recurrent connections (dropout rate 0.1). We trained using Adam, with a learning rate of 0.001 and a batch size of 100, a maximum of 50 epochs and early stopping. We also use label smoothing with a mixture weight $\epsilon = 0.2$, as described in \citet{szegedy2016}.

We ran all reported experiments at least three times and report the average of the computed metrics.
While the variance of the metrics varies between settings, it generally stays between $0.1$ and $0.25$ percent for $\text{Acc}_{\text{QM}}$.

\subsection{Results}
\label{sec:results}
We present our results, compared to previous and concurrent work in Table~\ref{tab:results}.
Our method compares well against previous works, achieving performance similar to Coarse2Fine~\cite{coarsefine} and close to MQAN~\citep{decathlon} which have more complicated architectures.
Approaches using execution-guided decoding (EG) show better performance at the expense of access to table content and repeated querying during decoding, and relies on the assumption that the query should not return empty result sets.
The concurrently developed oracle-based\footnote{We also investigated dynamic oracles in the previous unpublished version of this work from May 2018 (\url{https://openreview.net/forum?id=HJMoLws2z}).} approach of~\citet{incsql} improves upon our investigation of the oracle using the \texttt{ANYCOL} technique (see Related Work \af{section}). 
However, the \texttt{ANYCOL} technique assumes knowledge of table content during training and therefore might require retraining when table contents change.

In the following sections, we provide an ablation study, an in-depth analysis of the influence of the linearization order of query trees, as well as an error analysis.
The analysis reveals that the overall improvement in accuracy obtained from using the dynamic oracle can be attributed to improved prediction accuracy of WHERE clauses, which contain unordered elements.

\begin{table*}[htbp!]
\centering
\begin{tabu}{ l c c c  c c c }
\toprule
\multirow{2}{*}{} &
      \multicolumn{3}{c}{Dev Accuracies (\%)} &
      \multicolumn{3}{c}{Test Accuracies (\%)} \\
    & $\text{Acc}_{\text{LF}}$ & $\text{Acc}_{\text{QM}}$ &$\text{Acc}_{\text{EX}}$ & 	$\text{Acc}_{\text{LF}}$ & $\text{Acc}_{\text{QM}}$ & $\text{Acc}_{\text{EX}}$ \\
    \midrule
Seq2SQL (no RL)~\cite{seq2sql} & 48.2 & -- & 58.1 & 47.4 & -- & 57.1 \\
Seq2SQL (RL)~\cite{seq2sql} & 49.5 & -- & 60.8 & 48.3 & -- & 59.4 \\
Pointer-SQL~\cite{wang2018a} & 59.6 & -- & 65.2 & 59.5 & -- & 65.1 \\
*Seq2SQL (SQLNet) ~\cite{sqlnet} & 52.5 & 53.5 & 62.1 & 50.8 & 51.6 & 60.4 \\
SQLNet~\cite{sqlnet} & -- & 63.2 & 69.8 & -- & 61.3 & 68.0 \\
PT-MAML~\cite{huang2018} & 63.1 & -- & 68.3 & 62.8 & -- & 68.0 \\
TypeSQL~\cite{yu2018} & -- & 68.0 & 74.5 & -- & 66.7 & 73.5 \\
Coarse2Fine~\cite{coarsefine} & -- & -- & -- & -- & 71.7 & 78.5 \\
MQAN~\cite{decathlon} & -- & -- & -- & 72.4 & -- & 80.4 \\
\midrule
\multicolumn{7}{l}{\textit{(ours)}} \\
\ourname{} (\textit{shared softmax})& $70.2$ & $72.6$ & $79.0$ & $69.9$ & $72.1$ & $78.4$ \\ 
\ourname{} (\textit{point-or-generate}) & $70.0$ & $72.4$ & $78.5$ & $69.7$ & $71.7$ & $78.0$ \\ 
\ourname{} (\textit{shared softmax}) + oracle & -- & 73.4 & 79.4 & -- & 72.7 & 78.8 \\ 
\midrule
\multicolumn{7}{l}{\textit{(EG-based or concurrent work)}} \\
Pointer-SQL + EG(5)~\cite{wang2018b} & 67.5 & -- & 78.4 & 67.9 & -- &78.3 \\
Coarse2Fine + EG(5)~\cite{wang2018b} & 76.0 & -- & 84.0 & 75.4 & -- & 83.8 \\
IncSQL + oracle + \texttt{ANYCOL}~\cite{incsql} & 49.9 & -- & 84.0 & 49.9 & -- & 83.7 \\
IncSQL + oracle + \texttt{ANYCOL} + EG(5)~\cite{incsql} & 51.3 & -- & 87.2 & 51.1 & -- & 87.1 \\
\bottomrule
\end{tabu}
\caption{\label{tab:results} Evaluation results for our approach (middle section) and comparison with previously reported results (top part) and concurrent work or EG-based systems (bottom part). 
Note that \textit{*Seq2SQL} is the reimplementation of Seq2SQL~\cite{seq2sql} by SQLNet authors~\cite{sqlnet}. 
Some values in the table, indicated by ``--'', could not be filled because the authors did not report the metric or the metric was not applicable.
}
\end{table*}

\subsubsection{Ablation study}
\label{sec:results:ablation}

\af{Starting} from the best \af{variant of our} model 
\af{(i.e. the \textit{shared softmax} pointer-generator)
and standard TF based training, we want to investigate the 
the role of different model components and the different
training approaches.}

\begin{table*}[t]
\centering
\begin{tabu}{ l c c  c c }
\toprule
\multirow{2}{*}{} &
      \multicolumn{2}{c}{Dev Accs (\%)} &
      \multicolumn{2}{c}{Test Accs (\%)} \\
    & $\text{Acc}_{\text{LF}}$ & $\text{Acc}_{\text{QM}}$ & 	$\text{Acc}_{\text{LF}}$ & $\text{Acc}_{\text{QM}}$  \\
    \midrule
\ourname{} (\textit{shared softmax})& $70.2$ & $72.6$ & $69.9$ & $72.1$ \\ \midrule
$\cdot$ no constraints & $68.6$ & $70.9$ & $68.6$ & $70.5$ \\ 
$\cdot$ using constraints during training & $69.8$ & $72.2$ & $69.8$ & $71.9$ \\ $\cdot$ no label smoothing & $68.3$ & $70.5$ & $68.4$ & $70.1$ \\ 
\enspace $\cdot$ no label smoothing (\textit{point-or-generate}) & $68.7$ & $70.7$ & $68.5$ & $70.3$ \\ 
$\cdot$ no skip connections & $69.6$ & $72.0$ & $69.4$ & $71.6$ \\
\bottomrule
\end{tabu}
\caption{\label{tab:results:ablation} Performance of different variations of our approach.}
\end{table*}

Table~\ref{tab:results:ablation} presents the results of \af{this} ablation study.
Without constraints \af{enforcing the coherence of the decoded logical rule} at test time, the results drop by $1.6\%$.
While also using the constraints during training doesn't deteriorate results much, it results in slower training.

Label smoothing~\cite{szegedy2016} has a significant impact on performance. Label smoothing relaxes the target distribution and thus helps \af{to} reduce overfitting. While label smoothing improves the performance of both types of pointer-generators, 
\af{it brings more benefit for}
the \textit{shared softmax} version
 (2\% vs 1.4\% for \textit{point-or-generate}).

\af{Incorporating} skip connections \af{into the encoder and decoder of our model}
\af{improved} performance by $0.5\%$. This improvement is achieved because skip connections allow to bypass more complicated RNN encoding functions and  make predictions based on the more low-level features provided by word embeddings
\af{possible.}

\subsubsection{Influence of order in supervision}
\label{sec:results:order}
\begin{table}[b]
\centering
\begin{tabu}{ l c c   c c }
\toprule
\multirow{2}{*}{} &
      \multicolumn{2}{c}{Dev Accs (\%)} &
      \multicolumn{2}{c}{Test Accs (\%)} \\
    & $\text{Acc}_{\text{LF}}$ & $\text{Acc}_{\text{QM}}$ & 	$\text{Acc}_{\text{LF}}$ & $\text{Acc}_{\text{QM}}$\\
    \midrule
Original order (TF) & $70.2$ & $72.6$ & $69.9$ & $72.1$ \\ 
\midrule
$\cdot$ Reversed (TF) & -- & $72.6$ & -- & $72.1$ \\ 
$\cdot$ Arbitrary (TF) & -- & $70.4$ & -- & $69.6$ \\ 
$\cdot$ RL & $59.9$ & $71.4$ & $59.1$ & $70.6$ \\ 
$\cdot$ Oracle & $56.2$ & $73.4$  & $55.0$ & $72.7$ \\ 
\bottomrule
\end{tabu}
\caption{\label{tab:results:order} Results under different ways of linearizing the target query trees.
}
\end{table}
To investigate the influence of the order of the target sequences on the results,
we trained our model with teacher forcing and experimented with (1) reversing the original order of conditions in the WHERE clause and (2) training with target sequences where we assigned a different random order to the conditions in every trial.
The results indicate that the order of conditions in the linearization matters for the performance of TF based training to a small degree. Training with a randomly reassigned order of conditions in the WHERE clause results in a $2.5\%$ drop in test accuracy. 
However, reversing the order of conditions does not affect the results. 

\af{Furthermore, we trained our model with REINFORCE as well as with the dynamic oracle.}
\dl{In both methods, the originally provided order of the target sequence does not matter.}
Using REINFORCE (indicated by ``RL'' in Table~\ref{tab:results:order}) 
results in a $1.5\%$ drop in performance. 

The dynamic oracle as described in Alg.~\ref{alg:oracle} provides an improvement of $0.5\%$.
We can also see that $\text{Acc}_{LF}$ for the oracle is significantly lower compared to TF while $\text{Acc}_{QM}$ is on par with TF. 
Given that $\text{Acc}_{LF}$ is sensitive to the order of arbitrarily ordered clauses and $\text{Acc}_{QM}$ is not, 
this means that the oracle-trained models effectively learned to use alternative paths. 

Comparing the oracle to TF with arbitrarily reordered conditions in the WHERE clause shows that when the supervision sequences are not consistently ordered, training with TF can suffer. When training with the oracle, the order of unordered nodes as provided in supervision sequences does not matter. \dl{Thus, it can be beneficial (in this case by $3\%$ query accuracy) to use the oracle if the original linearization is arbitrary and can not be made consistent.}

\subsubsection{Error analysis}
\label{sec:results:errors}

\begin{table}[ht]
\centering
\begin{tabu}{ l r r}
\toprule
\multirow{2}{*}{} &
      \multicolumn{1}{c}{TF} &
      \multicolumn{1}{c}{Oracle} \\
\midrule
Whole Query & 27.4 \enspace \enspace & 26.6 \enspace \enspace \\ 
\enspace $\cdot$ SELECT & 53.0 \enspace \enspace & 54.6 \enspace \\
\enspace \enspace $\cdot$ Aggregator & 68.7 & 68.6 \\ 
\enspace \enspace $\cdot$ Column & 36.5 & 36.4  \\ 
\enspace \enspace $\cdot$ (Both) & 5.2 & 5.0  \\ 
\enspace $\cdot$ WHERE & 60.7 \enspace & 58.7 \enspace \\ 
\enspace $\cdot$ (Both) & 13.7 \enspace  & 13.3 \enspace   \\ 
\bottomrule\end{tabu}
\caption{
\label{tab:results:error} Errors for the TF-trained and oracle-trained models with \textit{shared softmax}. The reported numbers indicate percentages relative to the parent category (see text for details). Percentages are are derived from error counts averaged across different runs of the same setting.}
\end{table}

Table~\ref{tab:results:error} shows a breakdown of the causes of errors over the development set of \wikisql{}.
The numbers shown are percentages of error cases relative to the parent category.
For example, in 36.5\% of cases where the SELECT clause had an error, the column predicted in the SELECT clause was wrong.\footnote{Note that the percentages do not sum up to one since all components of a predicted clause can be wrong (which is indicated by the ``\textit{(Both)}''-labeled rows).}
From the presented numbers, we can conclude that the main cause of a wrongly predicted SELECT clause is an error in the predicted aggregator, while the main cause of error overall is the prediction of the WHERE clause.
Comparison of \af{ errors of a models trained with} TF \af{versus} Oracle reveals that oracle-trained models makes fewer mistakes in the WHERE clause, showing a 1\% improvement in WHERE clause accuracy, which is translated to the 0.5\% improvement in overall accuracy as observed in Table~\ref{tab:results}.

\section{Related Work}
\label{sec:related}
Earlier works on semantic parsing relied on CCG and other grammars~\cite{ccg2,berant2013}.
With the recent advances in recurrent neural networks and attention~\citep{bahdanau,see2017}, neural translation based approaches for semantic parsing have been developed~\cite{donglapata,nsm,rabinovich2017}. 

Labels provided for supervision in semantic parsing datasets can be given either as execution results or as an executable program (logical form).
Training semantic parsers on logical forms yields better results than having only the execution results~\cite{yih2016} but requires a more elaborate data collection scheme.
Significant research effort has been dedicated to train semantic parsers only with execution results. Using policy gradient methods (such as REINFORCE) is a common strategy~\cite{nsm,seq2sql}. Alternative methods~\cite{krishnamurthy2017,iyyer2017,guu2017} exist, which also maximize the likelihood of the execution results.

Similar to the \wikisql{} dataset that we used in our experiments are the \textsc{ATIS} and \textsc{WikiTableQuestions}~\cite{wikitableqa} datasets, which also focus on question answering over tables. In contrast to \wikisql{} however, both \textsc{ATIS} and \textsc{WikiTableQuestions} are significantly smaller and the latter does not provide logical forms for supervision and thus requires training with execution results as supervision~\cite{neuralprogrammer,haug2018,krishnamurthy2017}.
SQA~\cite{iyyer2017} is a dataset derived from \textsc{WikiTableQuestions} and focuses on question answering in a dialogue context. 

Previous works on \wikisql{}~\cite{seq2sql,sqlnet,huang2018,yu2018,wang2018a} generally incorporate both slot-filling and sequence decoding, predicting the SELECT clause arguments with separate slot-filling networks, and also include some form of a pointing mechanism. 
Seq2SQL~\cite{seq2sql} proposes an \textit{augmented pointer network} that also uses a pointer but encodes the question, column names and SQL tokens together, and completely relies on a pointer to generate the target sequence. To avoid the potential \textit{order-matters} problem, 
SQLNet~\cite{sqlnet} proposes a sequence-to-set model that makes a set inclusion prediction for the clauses in order to avoid decoding the conditions in any particular order. 
Both predict the SELECT clause arguments using separate specialized predictors. 
\citet{seq2sql} also use \citet{donglapata}'s \seqtoseq{} model as a baseline, however, get poor performance due to the lack of a pointer and column encoders.
\citet{yu2018} build upon SQLNet~\cite{sqlnet}'s slot filling approach, proposing several improvements such as weight sharing between SQLNet's subnetworks, and incorporate precomputed type information for question tokens in order to obtain a better question encoding. 
\citet{wang2018a} develop a model similar to ours; they propose a \seqtoseq{} model with copy actions. 
Similarly to \citet{seq2sql}, they encode the concatenation of column names and the question. Similarly to our work, they use a constrained decoder to generate SQL tokens or copy column names or question words from the encoded input sequence. In contrast to~\citet{wang2018a}, we encode column names separately, and independently from the question.
\citet{huang2018} experiment with meta-learning (MAML), using \citet{wang2018a}'s model.
Coarse2Fine~\cite{coarsefine} explores a middle ground between purely sequence and tree decoding models~\cite{alvarez2016,donglapata} and proposes a two-stage decoding process, where first a template (sketch) of the query is decoded and subsequently filled in.

Very recent and concurrent work on \wikisql{}~\citep{wang2018b,incsql} explores execution-guided (EG) decoding~\cite{wang2018b} and dynamic oracles~\cite{incsql}. 
Execution-guided decoding keeps a beam of partially decoded queries, which are filtered based on the execution, that is, if the partially decoded query can not be parsed, produces a runtime error or returns an empty result set (if we expect a non-empty output). 
However, this technique requires multiple queries to be executed against the database while decoding and thus is sensitive to the current state of the database.
A significant part of improvement obtained by EG decoding relies on the assumption that result sets should be non-empty.
IncSQL~\cite{incsql} also uses EG decoding, as well as a dynamic oracle extended with the \texttt{ANYCOL} technique, which adds the option to produce a wildcard column token that matches any column. During training, the wildcard column token is provided as an alternative to the true column token in the supervision sequence if it can be unambiguously resolved to the true column using the condition value. This makes the training process dependent on table contents and thus might result in a need to retrain when the table contents change.
IncSQL's model goes beyond ours by adding self- and cross-serial attention and a final inter-column BiLSTM encoder. In addition, they also feed column attention and question attention summaries as an input to the decoder.

\section{Conclusion}
\label{sec:conclusion}
In this work we present a \seqtoseq{} model adapted to the semantic parsing task of translating natural language questions to queries over tabular data. 
We investigated the order-matters problem, concluding that the order of conditions in the linearization of the query tree matters to small but significant degree for \wikisql{}.
In this context, we also evaluated the use of REINFORCE and a dynamic oracle for training the neural network-based semantic parser. Our experiments revealed that REINFORCE does not improve results and the oracle provides a small improvement, which can be attributed to improved decoding of the WHERE clause.
Furthermore, from the results we can conclude that training with a dynamic oracle is advisable if the original linearizations are inconsistently ordered.

\newpage
\bibliographystyle{aaai}

\end{document}